\documentclass[runningheads]{llncs}
\usepackage{graphicx}
\usepackage[dvipsnames]{xcolor}
\usepackage{booktabs}
\usepackage{multirow}
\usepackage[export]{adjustbox}

\usepackage{multirow}
\usepackage[export]{adjustbox}

\begin{document}
\title{BERT-Based Arabic Social Media \\ Author Profiling}

\author{Chiyu Zhang \and
Muhammad Abdul-Mageed}
\authorrunning{Zhang \& Abdul-Mageed}

\institute{Natural Language Processing Lab\\The University of British Columbia\\{\tt chiyuzh@mail.ubc.ca}, {\tt muhammad.mageeed@ubc.ca}}
\maketitle              
\begin{abstract}
We report our models for detecting age, language variety, and gender from social media data in the context of the Arabic author profiling and deception detection shared task (APDA)~\cite{rangel2019ADPA}. We build simple models based on pre-trained bidirectional encoders from transformers (BERT). We first fine-tune the pre-trained BERT model on each of the three datasets with shared task released data. Then we augment shared task data with in-house data for gender and dialect, showing the utility of augmenting training data. Our best models on the shared task test data are acquired with a majority voting of various BERT models trained under different data conditions. We acquire 54.72\% accuracy for age, 93.75\% for dialect, 81.67\% for gender, and 40.97\% joint accuracy across the three tasks.
\end{abstract}


\section{Introduction}
The proliferation of social media has made it possible to collect user data in unprecedented ways. These data can come in the form of usage and behavior (e.g., who likes what on Facebook), network (e.g., who follows a given user on Instagram), and content (e.g., what people post to Twitter). Availability of such data have made it possible to make discoveries about individuals and communities, mobilizing social and psychological research and employing natural language processing methods. In this work, we focus on predicting social media user age, dialect, and gender based on posted language. More specifically, we use the total of 100 tweets from each manually-labeled user to predict each of these attributes. Our dataset comes from the Arabic author profiling and deception detection shared task (APDA)~\cite{rangel2019ADPA}. We focus on building simple models using pre-trained bidirectional encoders from transformers (BERT)~\cite{devlin2018bert} under various data conditions. Our results show (1) the utility of augmenting training data, and (2) the benefit of using majority votes from our simple classifiers. 

In the rest of the paper, we introduce the dataset, followed by our experimental conditions and results. We then provide a literature review and conclude.



\section{Data}\label{data}
For the purpose of our experiments, we use data released by the APDA shared task organizers. The dataset is divided into train and test by organizers. The TRAIN set is distributed with labels for the three tasks of age, dialect, and gender. Following the standard shared tasks set up, the test set is distributed without labels and participants were expected to submit their predictions on test. The shared task predictions are expected by organizers at the level of users. The distribution has 100 tweets for each user, and so each tweet is distributed with a corresponding user id. As such, in total, the distributed training data has 2,250 users, contributing a total of 225,000 tweets. The official task test set contains 720,00 tweets posted by 720 users. For our experiments, we split the training data released by organizers into 90\% TRAIN set (202,500 tweets from 2,025 users) and 10\% DEV set (22,500 tweets from 225 users). The \textit{age} task labels come from the tagset \{\textit{under-25}, \textit{between-25 and 34}, \textit{above-35}\}. For dialects, the data are labeled with 15 classes, from the set \{\textit{Algeria}, \textit{Egypt}, \textit{Iraq}, \textit{Kuwait}, \textit{Lebanon-Syria}, \textit{Lybia}, \textit{Morocco}, \textit{Oman}, \textit{Palestine-Jordan}, \textit{Qatar}, \textit{Saudi Arabia}, \textit{Sudan}, \textit{Tunisia}, \textit{UAE}, \textit{Yemen}\}. The \textit{gender} task involves binary labels from the set \{\textit{male}, \textit{female}\}. 
\section{Experiments}\label{sec:exp}
As explained earlier, the shared task is set up at the user level where the age, dialect, and gender of each \textit{user} are the required predictions. In our experiments, we first model the task at the \textit{tweet} level and then port these predictions at the user level. For our core modelling, we fine-tune BERT on the shared task data. We also introduce an additional in-house dataset labeled with dialect and gender tags to the task as we will explain below. As a baseline, we use a small gated recurrent units (GRU) model. We now introduce our tweet-level models.

\subsection{Tweet-Level Models}
\subsubsection{Baseline GRU.} Our baseline is a GRU network for each of the three tasks. We use the same network architecture across the 3 tasks. For each network, the network contains a layer unidirectional GRU, with 500 units and an output linear layer. The network is trained end-to-end. Our input embedding layer is initialized with a standard normal distribution, with $\mu=0$, and $\sigma=1$, i.e., $W \sim N(0,1)$. We use a maximum sequence length of 50 tokens, and choose an arbitrary vocabulary size of 100,000 types, where we use the 100,000 most frequent words in TRAIN. To avoid over-fitting, we use dropout~\cite{srivastava2014dropout} with a rate of 0.5 on the hidden layer. For the training, we use the Adam~\cite{kingma2014adam} optimizer with a fixed learning rate of $1e-3$. We employ batch training with a batch size of 32 for this model. We train the network for 15 epochs and save the model at the end of each epoch, choosing the model that performs highest accuracy on DEV as our best model. We present our best result on DEV in Table~\ref{tab:res}. We report all our results using accuracy. Our best model obtains 42.48\% for age, 37.50\% for dialect, and 57.81\% for gender.  All models obtain best results with 2 epochs.

\subsubsection{BERT.} For each task, we fine-tune on the BERT-Base Muultilingual Cased model relesed by the authors~\cite{devlin2018bert}~\footnote{\url{https://github.com/google-research/bert/blob/master/multilingual.md}}. The model was pre-trained on Wikipedia of 104 languages (including Arabic) with 12 layer, 768 hidden units each, 12 attention heads, and has 110M parameters in entire model. The vocabulary of the model is 119,547 shared WordPices. We fine-tune the model with maximum sequence length of 50 tokens and a batch size of 32. We set the learning rate to $2e-5$ and train for 15 epochs. We use the same network architecture and parameters across the 3 tasks. As Table~\ref{tab:res} shows, comparing with GRU, BERT is 3.16\% better for age, 4.85\% better for dialect, and 2.45\% higher for gender. 
\subsubsection{Data Augmentation.} To further improve the performance of our models, we introduce in-house labeled data that we use to fine-tune BERT. For the gender classification task, we manually label an in-house dataset of 1,100 users with gender tags, including 550 \textit{female} users, 550 \textit{male} users. We obtain 162,829 tweets by crawling the 1,100 users' timelines. We combine this new gender dataset with the gender TRAIN data (from shared task) to obtain an extended dataset, to which we refer as \texttt{EXTENDED\_Gender}. For the dialect identification task, we randomly sample 20,000 tweets for each class from an in-house dataset gold labeled with the same 15 classes as the shared task. In this way, we obtain 298,929 tweets (\textit{Sudan} only has 18,929 tweets). We combine this new dialect data with the shared task dialect TRAIN data to form \texttt{EXTENDED\_Dialect}. For both the dialect and gender tasks, we fine-tune BERT on \texttt{EXTENDED\_Dialect} and \texttt{EXTENDED\_Gender} independently and report performance on DEV. We refer to this iteration of experiments as BERT\_EXT. As Table~\ref{tab:res} shows, BERT\_EXT is 2.18\% better than BERT for dialect and 0.75\% better than BERT for gender.~\footnote{We note that it was not possible for us to use external age-labeled data and hence we do not report on the age task with this data augmentation setting.}
\begin{table}[h!]
\centering
\caption{Tweet level results on DEV} \label{tab:res}
\footnotesize
\begin{tabular}{@{}lccc@{}}
\toprule
\multicolumn{1}{c}{} & \textbf{Age} & \textbf{Dialect} & \textbf{Gender}  \\ \midrule
\textbf{GRU}         & 42.48        & 37.50         & 57.81               \\
\textbf{BERT}        & 45.64        & 42.35      & 60.26                    \\
\textbf{BERT\_EXT}   & -            & 44.53      & 61.01                     \\ \bottomrule
\end{tabular}

\end{table}

\subsection{User-Level Models}
Our afore-mentioned models identify user's profiling on the tweet-level, rather than directly detecting the labels of a user. Hence, we follow the work of Zhang \& Abdul-Mageed~\cite{zhang2019no} to identify user-level labels. For each of the three tasks, we use tweet-level predicted labels (and associated softmax values) as a proxy for user-level labels. For each predicted label, we use the softmax value as a threshold for including only highest confidently predicted tweets. Since in some cases softmax values can be low, we try all values between 0.00 and 0.99 to take a softmax-based majority class as the user-level predicted label, fine-tuning on our DEV set. Using this method, we acquire the following results at the user level: \texttt{BERT} models obtain an accuracy of 55.56\% for age, 96.00\% for dialect, and 80.00\% for gender. \texttt{BERT\_EXT} models achieve 95.56\% accuracy for dialect and 84.00\% accuracy for gender.

\subsection{APDA@FIRE2019 submission}
\textbf{First submission.} For the shared task submission, we use the predictions of BERT\_EXT as out first submission for gender and dialect, but only BERT for age (since we have no BERT\_EXT models for age, as explained earlier). In each case, we acquire results at tweet-level first, then port the labels at the user-level as explained in the previous section. For our second and third submitted models, we also follow this method of going from tweet to user level. 
\textbf{Second submission.} We combine our DEV data with our EXTENDED\_Dialect and EXTENDED\_Gender data, for dialect and gender respectively, and train our second submssions for the two tasks. For age second submsision, we concatenate DEV data to TRAIN and fine-tune the BERT model. We refer to the settings for our second submission models collectively as BERT\_EXT+DEV.

\textbf{Third submission.} Finally, for our third submission, we use a majority vote of (1) first submission, (2) second submission, and (3) predictions from our user-level BERT model. These majority class models (i.e., our third submission) achieve best results on the official test data. We acquire 54.72\% accuracy for age, 81.67\% accuracy for gender, 93.75\% accuracy for dialect, and 40.97\% joint accuracy.

\begin{table}[h!]
\centering
\caption{Results of our submissions on official test data (user level)}
\begin{tabular}{@{}l|lcccc@{}}
\toprule
             & \textbf{Exp. Condition} & \textbf{Age}   & \textbf{Dialect} & \textbf{Gender}  & \textbf{Joint} \\ \midrule
\textbf{Submission 1} & \textbf{BERT\_EXT}     & 54.72     & 93.33   & 77.08    & 38.75 \\
\textbf{Submission 2} & \textbf{BERT\_EXT+DEV} & 54.72      & 92.64   & 81.67   & 40.97 \\
\textbf{Submission 3} & \textbf{MAJ\_CLASS}     & 54.72     & 93.75   & 81.67    & 40.97 \\ \bottomrule
\end{tabular}

\end{table}

\section{Related Works}\label{sec:rel}

\textbf{Arabic.} \textit{Arabic} is a term that refers to a collection of languages, varieties, and dialects. The standard variety, Modern Standard Arabic (MSA), is the one usually used in formal communication and educational settings. Arabic also has a wide range of under-studied varieties and dialects that classically used to be categorized in a coarse-grained fashion (e.g., Levantine, North African)~\cite{habash2010introduction,versteegh2014arabic,mageed2015subjectivity,elaraby2018deep,abdul2017modeling}. More recent treatments focus on fine-grained categorizations such as country and city levels~\cite{mubarak2014using,sadat2014automatic,mageed2018city,salameh2018fine,qwaider2018shami,zhang2019no}. Differences between varieties of Arabic happen at various linguistic levels, including including phonological, morphological, lexical, and syntactic~\cite{holes2004modern,bassiouney2009arabic,palva2006dialects,mageed2015subjectivity}.

\textbf{Social Media Author Profiling.} \textit{Author profiling} is the term usually used to refer to detecting a host of attributes of (often social media) users. This include identifying attributes such as age, gender, educational level, economic class, stance or ideology~\cite{colleoni2014echo,preoctiuc2017beyond}, personality~\cite{schwartz2013personality,matz2017psychological,bleidorn2018using,hinds2019human}, moral traitstraits~\cite{johnson2018classification,pang2019language}, and other socilogical and psychological constructs. Author profiling based on text~\cite{gamon2004linguistic,argamon2019register} is rooted in computational stylometry~\cite{goswami2009stylometric,verhoeven2018two} and has traces in the early work of Holmes~\cite{holmes1998evolution}. The task of author profiling has also been approached from network perspective where cues based on friending, following, mentioning, and commenting have been leveraged for identifying author attributes~\cite{mitrou2014social} for author profiling. In addition, the PAN author profiling shared task~\cite{rangel2013overview,rangel2014overview,rangel2015overview,rangel2016overview,rangel2017overview} was established to advance related work. More information about PAN can be found in~\cite{potthast2019decade}.

%
\textbf{Age and Gender.} 
A number of studies have been conducted on English-based age and gender detection, including ~\cite{rao2010classifying,flekova2016exploring,burger2011discriminating,verhoeven2016twisty,daneshvar2018gender}. Many of these works use feature engineering such as text n-gram and topic models~\cite{schwartz2013personality,sap2014developing}. In these works, age is either cast as a multi-class classification task with, e.g., labels from the set \{\textit{10-19, 20-29, 30-39}\} or as a regression task~\cite{nguyen2011author}. Other works model age with both classification and regression combined~\cite{chen2019joint}. 
With rare exceptions~\cite{rangel2017overview,alrifai2017arabic}, we do not know of work on Arabic targeting age and gender.









\section{Conclusion}\label{conc}
In this work, we described our submitted models to the Arabic author profiling and deception detection shared task (APDA)~\cite{rangel2019ADPA}. We focused on detecting age, dialect, and gender using BERT models under various data conditions, showing the utility of additional, in-house data on the task. We also showed that a majority vote of our models trained under different conditions outperforms single models on the official evaluation. In the future, we will investigate automatically extending training data for these tasks as well as better representation learning methods.  
\section{Acknowledgement}
We acknowledge the support of the Natural Sciences and Engineering Research Council of Canada (NSERC), the Social Sciences Research Council of Canada (SSHRC), and Compute Canada (\url{www.computecanada.ca}).

%
%
%
\bibliographystyle{splncs04}
\bibliography{llncs.bib}
\end{document}